# CarDiff: Anatomically Aware Dental Caries Synthesis using Diffusion

**Sanyam Jain[1*], Bruna Neves de Freitas[2], Alexandros Iosifidis[3], Ruben Pauwels[1]**

[1]Department of Dentistry and Oral Health, Aarhus University, 8000 Aarhus, Denmark;
[2]Aarhus Institute of Advanced Studies, Aarhus University, 8000 Aarhus, Denmark;
[3]Faculty of Information Technology and Communication Sciences, Tampere University, Finland

*** Correspondence:**
Corresponding Author
sanyam.jain@dent.au.dk

## Abstract

Synthetic medical image generation can help address limited annotated data in dental imaging, particularly for under-represented caries classes. In this work, we present CarDiff, a semi-supervised framework for dental caries synthesis in panoramic radiographs. The method combines paired mask-image learning with self-supervised generation from augmented masks, while incorporating anatomy-aware conditioning between lesion regions and surrounding structures. Semantic consistency is enforced through segmentation-guided supervision and adversarial anatomical alignment. Experiments on the MLUA-DC1000 dataset show that CarDiff achieves improved synthesis quality compared with existing semantic synthesis baselines, obtaining the best Fréchet Inception Distance (FID) (90.76) and Learned Perceptual Image Patch Similarity(0.48) scores among evaluated methods. We further assess the generated radiographs for downstream segmentation augmentation across seven architectures. Using a 5× augmented training set improves average Dice score by up to 6.5%, with larger gains observed for medium and combined lesion classes. These results suggest that anatomically consistent diffusion synthesis may be useful for dental radiograph augmentation under limited supervision. The code is available at https://github.com/s4nyam/CarDiff.

## 1 Introduction

Recent advances in generative modeling have enabled semantic image synthesis for medical imaging tasks, including synthetic radiography and automatic segmentation, but many approaches primarily learn pixel-space correlations without modeling anatomical structure or pathology relationships (Locatello et al., 2020; Pinto-Coelho, 2023). In dental radiography, limited annotated data and under-represented lesion classes remain major challenges for robust caries analysis. To address this, we propose CarDiff, a semi-supervised framework for dental caries synthesis in panoramic radiographs that combines paired mask-image learning inspired by self-supervised representation learning (Assran et al., 2023). The framework incorporates anatomy-aware conditioning between lesion regions and surrounding structures, while enforcing semantic consistency and anatomical plausibility during synthesis. We further evaluate the generated radiographs for downstream segmentation augmentation across seven architectures, demonstrating consistent improvements for several lesion categories under limited supervision.

## 2 Related Work

Recent semantic image synthesis methods use different conditioning strategies for controllable generation. SPADE (Park et al., 2019) introduced spatially adaptive normalization for Generative Adversarial Networks (GAN) based synthesis, while SISDM (Wang et al., 2022) extended semantic conditioning to diffusion models. SCDM (Ko et al., 2024) improved robustness through stochastic perturbation of semantic maps, and SegDiff (Konz et al., 2024) demonstrated segmentation-guided synthesis for anatomically consistent medical image generation. Although generative models have been explored for synthetic radiography and segmentation tasks, many approaches primarily learn pixel-level correlations without considering anatomical structure or pathology relationships (Pinto-Coelho, 2023). Motivated by these limitations, we investigate a semi-supervised diffusion framework for dental caries synthesis in panoramic radiographs.

## 3 Methodology

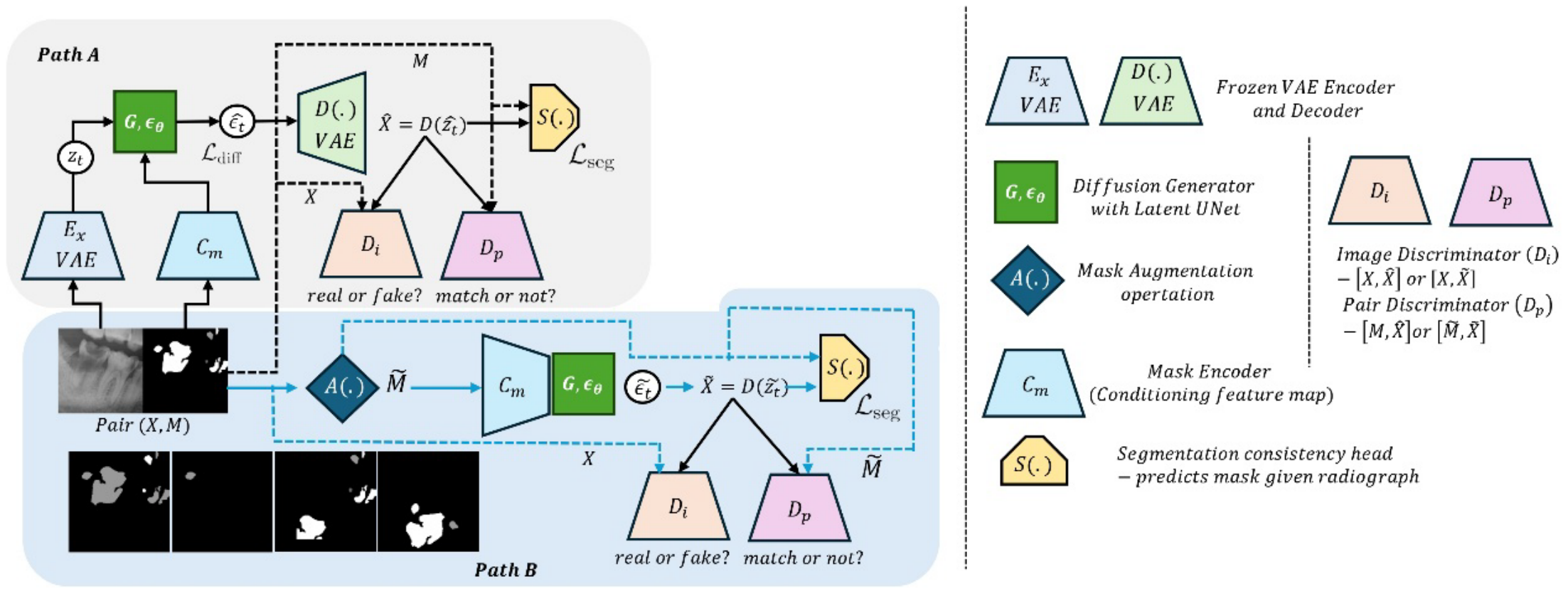

**Figure 1** *Overview of the CarDiff framework. The model performs semi-supervised segmentation-guided dental radiograph synthesis through two complementary training paths. Path A learns from paired mask-image samples using a latent diffusion generator, while Path B synthesizes new radiographs from augmented masks in a self-supervised manner. A segmentation consistency head and adversarial discriminators enforce semantic alignment, radiographic realism, and mask-image correspondence during training.*

*CarDiff* is a semi-supervised segmentation-guided diffusion framework for dental caries synthesis in panoramic radiographs. The framework combines paired mask-image learning with self-supervised generation from augmented segmentation masks to maximize learning under limited supervision. In the paired training path, real radiograph-mask pairs are encoded into a latent space using a frozen VAE encoder-decoder, and a latent diffusion generator learns to reconstruct denoised latent representations conditioned on segmentation mask embeddings. A segmentation consistency head re-segments synthesized radiographs and compares them with the conditioning masks using Dice supervision to preserve lesion geometry and semantic alignment. To further improve data diversity, a self-supervised branch generates new anatomically plausible masks using dilation, erosion, elastic deformation, and CutMix transformations, which are then used to synthesize additional radiographs without requiring paired real images. Adversarial discriminators promote radiographic realism and mask-image correspondence during training. In addition, contextual conditioning using patch-based feature interactions and global attention is integrated into the denoising U-Net through FiLM modulation (Perez et al., 2018) to provide anatomical consistency during synthesis. Together, these components enable anatomically consistent and semantically faithful radiograph generation for downstream segmentation augmentation under limited annotated data settings. The approach is shown in Figure 1.

# 4 Results and Discussion

The MLUA-DC1000 (Wang et al., 2023) dataset was used for training, validation, and testing, consisting of 2615 panoramic radiograph regions of interest extracted from 600 annotated radiographs. We first evaluated synthesis quality by comparing *CarDiff* with SPADE, SISDM, SCDM, and SegDiff using FID, Kernel Inception Distance (KID), Inception Score (IS), and LPIPS metrics. As shown in Figure 2 (left), *CarDiff* achieved the best FID (90.76) and LPIPS (0.48) among all evaluated methods, improving over the next-best baseline (SCDM) by approximately 5% and 6.6%, respectively. Although SCDM obtained slightly better KID and IS values, qualitative comparisons showed that *CarDiff* generated radiographs. In contrast, SegDiff and SPADE showed reduced visual quality and poorer metric performance across most evaluation measures. The synthesized samples are shown in Figure 2 (right).

| Model | FID ↓ | KID ↓ | IS ↑ | LPIPS ↓ |
|---|---|---|---|---|
| **CarDiff** | **90.76** | 0.021 | 2.65 ± 0.27 | **0.480** |
| SCDM | 95.55 | **0.016** | **2.75 ± 0.28** | 0.514 |
| SISDM | 99.96 | 0.024 | 2.40 ± 0.18 | 0.513 |
| SegDiff | 126.72 | 0.071 | 2.21 ± 0.18 | 0.636 |
| SPADE | 164.05 | 0.091 | 2.18 ± 0.23 | 0.517 |

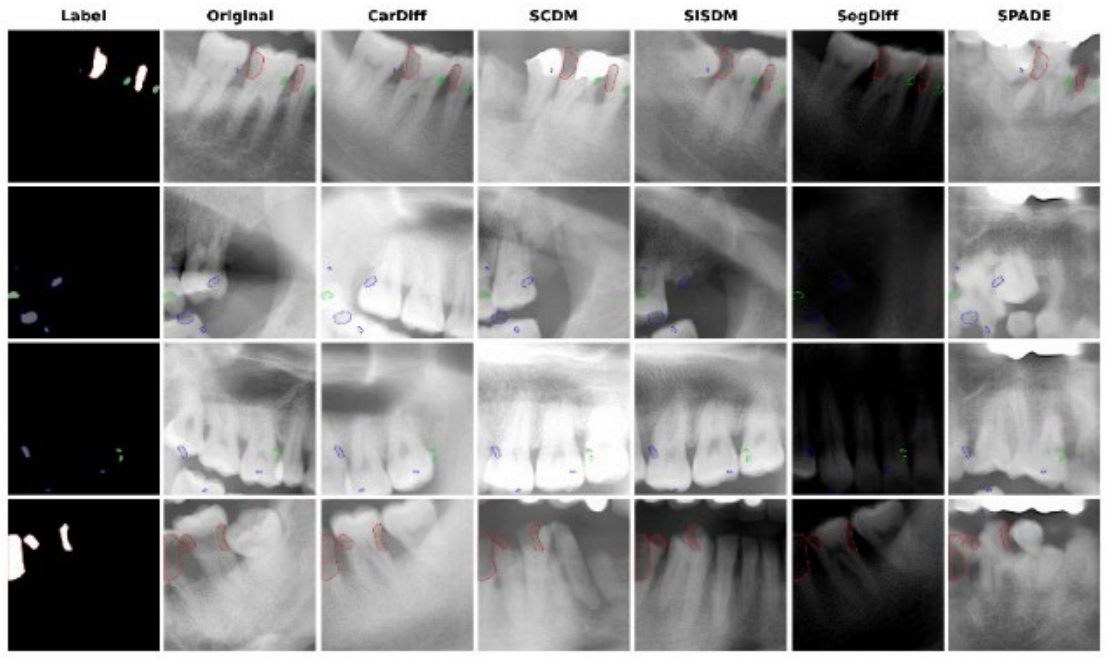

**Figure 2** *(Left)* Quantitative comparison of different models using FID, KID, IS, and LPIPS. Arrows indicate whether lower ↓ or higher ↑ values reflect better performance. *(Right)* Synthesized ROIs compared with original label-image pairs for CarDiff, SPADE, SCDM, SISDM and SegDiff. The images are overlapped with boundaries in color: Blue for superficial caries, Green for medium caries, and Red for deep caries.

We further evaluated whether the synthesized radiographs improved downstream segmentation performance through data augmentation. Using *CarDiff*, we generated 13075 synthetic radiographs conditioned on processed segmentation masks and evaluated augmentation performance across seven segmentation architectures (Iakubovskii, 2019), including U-Net, U-Net++, FPN, LinkNet, MA-Net, PAN, and PSPNet. As summarized in Table 2, augmentation with synthetic data generally improved Dice and Intersection over Union (IoU) scores across most architectures and lesion categories. The largest improvement was observed for medium caries (MC) using U-Net++, where Dice score increased from 0.23 to 0.32, corresponding to a 39% relative improvement. Combined lesion categories such as MC+SC also showed consistent gains across multiple architectures. Performance improvements for superficial caries (SC) were comparatively smaller and occasionally decreased for certain models, indicating that synthesis of subtle lesion patterns remains challenging. Overall, the results suggest that segmentation-guided diffusion synthesis can provide useful augmentation for dental caries analysis under limited annotated data settings.

# 5 Funding and Acknowledgements

This work was funded by the Independent Research Fund Denmark, project "Synthetic Dental Radiography using Generative Artificial Intelligence", grant ID 10.46540/3165-00237B. This work utilized computational resources from the LUMI supercomputer, provided by the EuroHPC Joint Undertaking, during the period from 10 October 2025 to 30 January 2026 under project ID DeiC-AU-S5-202500020 provided by the Danish e-Infrastructure Cooperation. Authors declare no conflicts.